# AN INITIAL STUDY ON ESTIMATING AREA OF A LEAF USING IMAGE PROCESSING

G.D. Illeperuma
Department of Physics
The Open University of Sri Lanka
Nawala, Nugegoda, Sri Lanka
gdilleperuma@gmail.com

*ABSTRACT*

Calculating leaf area is very important. Computer aided image processing can make this faster and more accurate. This include scanning the leaf, converting it to binary image and calculation of number of pixels covered. Later this is converted to $mm^2$.

## I. INTRODUCTION

Leaf area meters are used to estimate the surface area of a leaf which is useful in variety of fields including agronomy, ecology, plant physiology, plant pathology, carbon cycle studies and entomology [1] [2]. Manual method of measuring the surface area of a leaf includes drawing the outline of the leaf in a grid paper and counting squares covered by the leaf. This is time consuming and labor intensive. When a square is partially covered by the leaf including that square is subjective [1] [3]. Therefore, same leaf measured by two persons can lead to different results. Another problem is the accuracy. Minimum possible measurable area is limited by the size of the square in the grid. It is impractical to have a square smaller than 1 $mm^2$ which limits the accuracy. Electronic measuring equipment's are available as a solution to above problems. But these are expensive and are not readily available to the majority.

In this research an alternative low cost method is proposed to measure the leaf area and length of the leaf. It is based on image processing and only requires a scanner and a computer.

## II. METHODOLOGY

This process consists of several steps. (Figure 1)

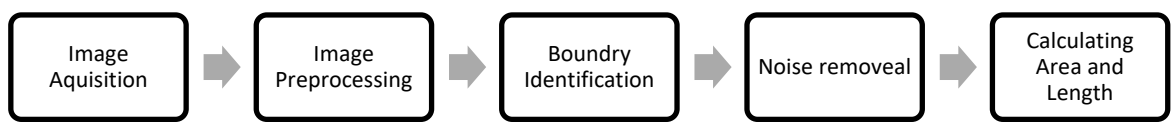

Figure 1. Steps in calculating leaf area

**IMAGE ACQUISITION:**

First step is to capture the image and digitize it. High resolution images provide more accurate results. It is also desired to have a uniform background to distinguish leaf area from the background. All these conditions are satisfied by a scanner.

Acquisition of the image is carried out by scanning the leaf using a scanner. It is important to make sure that leaf is flat on the scanner. Some scanners have software interpolation to increase the resolution. This can blur the edges of the leaf which can leads to inaccurate results. By setting the scanning resolution to the true resolution of the charge coupled device (CCD) this can be avoided.

Region of leaf is identified using the brightness differences of the leaf and the background. Any color background can be used as long as it has a different color than the leaf. For majority of cases a white background is suitable. In rare cases where leaf itself has a light color, black background can be used. Finally, image is loaded to the computer program which was developed to calculate the area.

**IMAGE PRE-PROCESSING:**

User is allowed to select a rectangular region which contains the leaf. This reduces the number of pixels needed to be processed in later stages thus increasing the efficiency. Then the image is converted to grayscale (Figure 2).

**LEAF BOUNDARY IDENTIFICATIONS:**

At the options panel user can select the background color. If the back ground is white, algorithm would determine darker regions as part of the leaf and vice versa [4]. Using a slider user can select the brightness threshold for the separation. A preview is provided which displays the currently selected region with an overlap image of leaf. This along with zoom functionality allows user to correctly set the brightness.

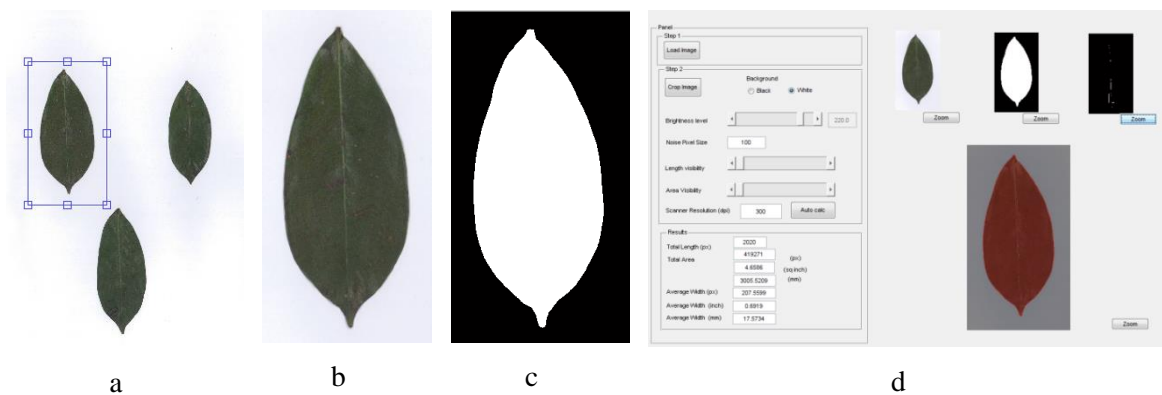

Figure 2. (a) Crop selection (b) Cropped image (c) Identified leaf area (d) User interface

**NOISE REMOVAL:**

Noise can appear on the image as a result of thermal noise in the CCD or due to dust particles on the scanner. Removing such noises increases the accuracy. An assumption is made that noise regions are considerably smaller than the leaf area. First all connected pixels are identified and the number of pixels in each object is calculated. If the area of an object is smaller than the value specified by the user, that object is removed [5].

**CALCULATION OF SURFACE AREA AND LENGTH IN PIXELS:**

Leaf area is calculated by counting the number of pixels in leaf region. To calculate the length of the leaf, image is thinned. This morphological operation results in a line which has a width of one pixel. Therefore, number of pixels in this image is equal to the length of the line [6]. Average width is calculated by dividing the total area by the length of the leaf.

**CONVERTING PIXEL READINGS TO METRIC UNITS:**

Resolution of a scanner is the number of dots it has per inch (dpi).

$$\text{Resolution (dpi)} = \frac{\text{Number of dots in length x in inches}}{\text{length x in inches}}$$

Real length and area can be calculated in inches by above formula. Then these values are converted to metric units.

An alternative method is provided when the resolution is unknown. While scanning, user need to place an object with a known length such as a ruler as a reference. By pressing the 'auto calc' button in resolution panel user is provided with the scanned image with the reference object. By selecting the two points of the reference object and entering the real distance between those two points, resolution is automatically calculated.

### III. RESULTS AND DISCUSSION

Theoretically error in this method is only depended on the resolution of the scanner. But some of the leaves distorted their shape while flatten on the scanner surface and some scanners did not provide the mentioned resolution which introduced errors in the result.

Using 300 dpi scanner and reference objects it was calculated error of the area measured using image processing method is within the 5%. This can be improved using high resolution scanners.

It is noted that some leaves contain different color regions. It may be useful to calculate area of these regions separately. To achieve such, color image is converted to HSV color space where hue component corresponds to the color. By providing a range for the hue it is possible to select a region with approximately similar color. After the selection of the region, same steps can be followed as mention above, to determine the area of the region.

It is concluded using scanner and image processing techniques that it is possible to calculate the area of a leaf. Although some sources of errors were identified their effect is minimal and can be neglected in practical situations.

IV. REFERENCES

[1] S. Yoshida, D. A. Forni, J. H. Cock, and K. A. Gomez, "Measurement of leaf area, leaf area index, and leaf thickness," *Lab. Man. Physiol. Stud. rice*, pp. 69–72, 1976.

[2] H. Fang, Z. Xiao, Y. Qu, and J. Song, "Leaf Area Index," in *Advanced Remote Sensing*, 2012, pp. 347–381.

[3] W. W. Wilhelm, K. Ruwe, and M. R. Schlemmer, "Comparison fo tree leaf area index meter ins corn canopy," *Crop Sci.*, vol. 40, pp. 1179–1183, 2000.

[4] P. S. Hiremath and J. Pujari, "Content Based Image Retrieval based on Color , Texture and Shape features using Image and its complement," *Int. J. Comput. Sci. Secur.*, vol. 1, pp. 25–35, 2007.

[5] A. Koschan and M. Abidi, *Digital Color Image Processing*. 2008.

[6] C. Vision and I. P. Toolbox, "Edge and Corner Detection," *Computer (Long. Beach. Calif).*, vol. 27, no. cmd, pp. 1–3, 2005.